\documentclass[letterpaper]{article} 
\usepackage{aaai24}  
\usepackage{times}  
\usepackage{helvet}  
\usepackage{courier}  
\usepackage[hyphens]{url}  
\usepackage{graphicx} 
\usepackage{natbib}  
\usepackage{caption} 
\usepackage{algorithm}
\usepackage{algorithmic}
\usepackage{bm}
\usepackage{algorithm}
\usepackage{algorithmic}
\usepackage{multirow}
\usepackage{booktabs}
\usepackage{latexsym}
\usepackage{caption}
\usepackage{subcaption}
\usepackage{amsmath}
\usepackage{algorithm}
\usepackage{algorithmic}
\usepackage{extarrows}
\usepackage{amssymb}
\usepackage{multicol}
\usepackage{multirow}
\usepackage{amsmath}
\usepackage{color}
\usepackage{mathrsfs}
\usepackage{subcaption}
\usepackage{tabularx}
\usepackage{xcolor}
\usepackage{epstopdf}
\usepackage{graphicx}
\usepackage{newfloat}
\usepackage{listings}
\DeclareCaptionStyle{ruled}{labelfont=normalfont,labelsep=colon,strut=off} 
\floatstyle{ruled}
\newfloat{listing}{tb}{lst}{}
\floatname{listing}{Listing}
\title{DA-Net: A Disentangled and Adaptive Network for Multi-Source Cross-Lingual Transfer Learning
}
\author{
   Ling Ge\textsuperscript{\rm 1},
    Chunming Hu\textsuperscript{\rm 1,\rm2,\rm3,}\thanks{Corresponding authors.},
    Guanghui Ma\textsuperscript{\rm1},
    Jihong Liu\textsuperscript{\rm4,}\footnotemark[1],
    Hong Zhang\textsuperscript{\rm5},
}
\affiliations{
    \textsuperscript{\rm 1} School of Computer Science and Engineering, Beihang University, Beijing, China \\
    \textsuperscript{\rm 2} College of Software, Beihang University, Beijing, China \\
        \textsuperscript{\rm 3} Zhongguancun Laboratory, Beijing, China \\
    \textsuperscript{\rm 4}School of Mechanical Engineering and Automation, Beihang University, Beijing, China\\
        \textsuperscript{\rm 5}National Computer Network Emergency Response Technical Team / Coordination Center of China, Beijing, China\\
    \{geling, hucm, maguanghui, ryukeiko\}@buaa.edu.cn, zhangh@isc.org.cn
}

\usepackage{bibentry}

\begin{document}

\maketitle

\begin{abstract}
Multi-Source cross-lingual transfer learning deals with the transfer of task knowledge from multiple labelled source languages to an unlabeled target language under the language shift. Existing methods typically focus on weighting the predictions produced by language-specific classifiers of different sources that follow a shared encoder. However, all source languages share the same encoder, which is updated by all these languages. The extracted representations inevitably contain different source languages' information, which may disturb the learning of the language-specific classifiers. Additionally,  due to the language gap,  language-specific classifiers trained with source labels are unable to make accurate predictions for the target language. Both facts impair the model's performance. To address these challenges, we propose a Disentangled and Adaptive Network ~(DA-Net). Firstly, we devise a feedback-guided collaborative disentanglement method that seeks to purify input representations of classifiers, thereby mitigating mutual interference from multiple sources. Secondly, we propose a class-aware parallel adaptation method that aligns class-level distributions for each source-target language pair, thereby alleviating the language pairs' language gap. Experimental results on three different tasks involving 38  languages validate the effectiveness of our approach.
\end{abstract}
\section{Introduction}

Recent advances in multilingual models~\cite{DBLP:conf/naacl/DevlinCLT19,DBLP:conf/nips/ConneauL19} have enabled significant improvements in many cross-lingual tasks,  owing partly  to the availability of large-scale annotated data.
However, some   tasks in no-annotated languages may suffer from the so-called “data
hungriness” and fail to enjoy  this technological advancement.
This problem can be alleviated by conducting cross-lingual
transfer learning 
~\cite{DBLP:conf/emnlp/ZengJYWLC22,DBLP:journals/tois/MoreoPS23,DBLP:journals/corr/abs-2305-09148}, which seeks  to transfer language-independent knowledge from the labelled  language (source) to the unlabeled language  (target)~\cite{DBLP:conf/acl/SherborneL22,DBLP:conf/coling/Ding0FM0ZLTC22,DBLP:journals/corr/abs-2301-08855}, so that unlabeled languages can benefit from existing techniques.


\begin{figure}[t]
    \centering
\includegraphics[width=0.45\textwidth]{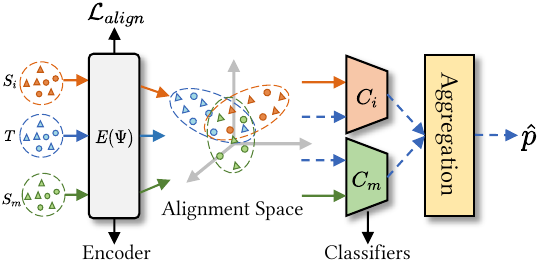}
\caption{Existing works align all languages through alignment loss~$\mathcal{L}_{align}$,  causing the model 
 only retaining  information invariant across all  languages. }
 \label{fig:introduction_fig}
\end{figure}

Despite impressive results,   most existing works  focus only on the single-source setting and fail to consider a more realistic scenario where multiple sources with different languages are available.
Since the target language may be similar to different source languages over various aspects, i.e., word order, capitalization, and script style~\cite{DBLP:conf/acl/HuJBWHHT20,DBLP:conf/acl/VriesWN22},  
multi-source cross-lingual transfer learning~\cite{DBLP:journals/corr/abs-1912-01389, DBLP:conf/acl/WuLKLH20}  generally produces superior performance~\cite{,DBLP:conf/acl/HuJBWHHT20} compared to  single-source algorithms and  has garnered much interest.

The competitive approaches~\cite{DBLP:conf/acl/ChenAHWC19,DBLP:conf/acl/JinD0LCDQ22} typically learn a shared  encoder, along with language-specific classifiers, which yield an ensemble prediction for  target language.
However, these approaches have  two  limitations. (1) Since multiple sources share one same  encoder, the   representations extracted by the  encoder inevitably contain information from different  source languages, which may confuse the optimisation of language-specific classifiers.
(2) Due to the language gap,  language-specific classifiers trained with source labels are unable to make accurate predictions for  target samples. Although these methods attempt to  alleviate this problem via performing   all languages alignment (shown in Fig \ref{fig:introduction_fig}), unfortunately, this constraint is so strict that the shared features in each source-target language pair would be eliminated, thus deteriorating the adaptability of all  language pairs and exacerbating this problem. Obviously, both  limitations  impair the model's performance.

In this paper, we propose a \textbf{D}isentangled and \textbf{A}daptive \textbf{Net}work (\textbf{DA-Net}) to tackle
the two problems simultaneously. The proposed architecture has several separate language branch networks  following a shared  encoder to extract  specific structures. Each language branch network consists of  a language-specific  disentangler, adaptor and classifier.

\begin{figure*}[t]
    \centering
  \includegraphics[width=0.90\textwidth]{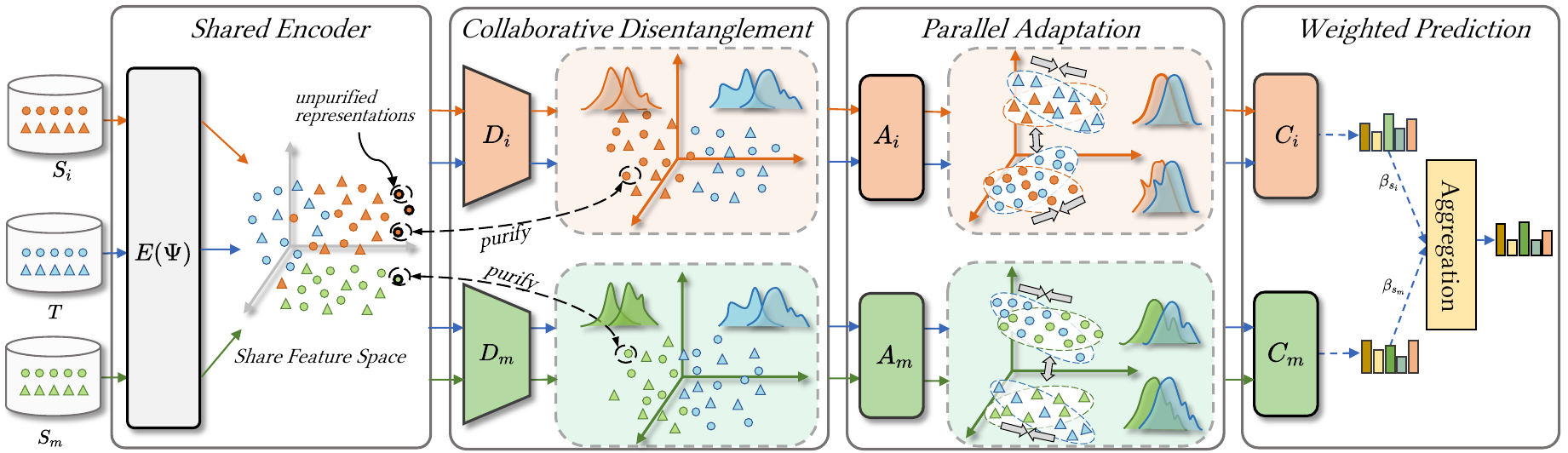}
    \caption{An illustration of our proposed model.}
    \label{fig:model}
\end{figure*}
Firstly, we propose a Feedback-guided Collaborative  Disentanglement (FCD) method, which utilizes supervised signals from different classifiers to guide disentanglers to purify the shared representations, thereby mitigating mutual interference from multiple sources.
We argue that a language-specific classifier can function as a detector to identify whether input representations contain the corresponding language information. 
The more information it contains about the related language, the better the language-specific classifier performs; otherwise, the classifier performs poorly.
Considering this,  we feed the representations generated by disentanglers into classifiers of their corresponding branch networks, and maximize the classification performance to preserve their respective language-specific information. Simultaneously, we feed these representations into other branches and minimize the classification performance to eliminate information irrelevant to the current source.

Secondly, we propose a Class-aware  Parallel Adaptation (CPA) method, which seeks to align  class-level  distributions  for   source-target language pairs at each  adaptor, thus bridging the 
language gap across languages. 
Class-level alignment can  force the model better to acquire the shared  semantics of class  and 
 avoid class-level feature mismatch issues  across languages ~\cite{DBLP:conf/emnlp/NguyenNMN21,DBLP:journals/corr/abs-2301-08855}, thereby enhancing the performance of language-specific classifiers for target samples.
To achieve this, we design a   distribution-based contrastive learning
to minimise the intra-class distribution distance and maximise the inter-class distribution distance across languages, thereby learning class discriminative shared features and ultimately improving language adaptation of  language pairs.

Finally, we conduct experiments  on three different tasks, including  Named Entity Recognition~(NER), Review Rating Classification~(RRC)  and Textual Entailment  Prediction~(TEP)  involving $38$ languages, and the results demonstrate the effectiveness of our DA-Net.

Summarily, we make the following contributions: (1) We propose the DA-Net  for multi-source cross-lingual tasks, which utilizes multiple source languages to enhance the model's generalisation  for the unlabeled target language. (2) We devise the FCD method to mitigate  mutual  interference from multiple  sources. (3) We propose the CPA method  to capture the shared class-level semantics of language pairs.

\section{Related Works}

To alleviate the data-scarcity issue for some languages, many cross-lingual transfer learning methods  have been developed   to learn  well adaptation models.  
Most  recent studies focus on the bilingual transfer case and  can be grouped into three categories: (1) Data-based  approaches  utilize machine translation and label projection\cite{DBLP:conf/emnlp/JainPL19,DBLP:conf/emnlp/YangHMY0ZGLW22} to create pseudo-training data for the target language. (2) Feature-based approaches rely on   features alignment to  diminish the language shift \cite{DBLP:conf/acl/ChenJW0G20,DBLP:journals/corr/abs-2301-08855}. (3) The  distillation-based methods \cite{DBLP:conf/ijcai/WuLKHL20,DBLP:conf/kdd/LiangGPSZZJ21,DBLP:conf/emnlp/MaCGLGL0L22} enable the student network to gain task knowledge from soft labels predicted by the teacher network on the target language. 
However, these studies are designed on the single-source assumption and
fail to deal with multiple source languages.

The Multi-source cross-lingual transfer is both feasible and valuable in practice  and has received increasing attention in application fields. 
Recent methods, driven by the  weighted combining rule, learn language-specific classifiers and obtain a weighted ensemble prediction for target samples. For instance, MulTS \cite{DBLP:conf/acl/ChenJW0G20} trains one specific  network for each source language separately  to derive the combined final prediction, which  causes a substantial computational burden.  MAN ~\cite{DBLP:conf/acl/ChenAHWC19}  and G-MOE ~\cite{DBLP:conf/acl/JinD0LCDQ22} force multiple source languages to share one same encoder, avoiding the huge parameter computation problem. They all perform  alignment among multiple sources and target  to extract language-invariant  knowledge. However, this rigorous constraint results in too little information the encoder learns.
In addition, none  of these approaches considers the language interference issue associated with   the shared encoder.

\section{Methodology}
In  multi-source  scenario, we consider $M$ labeled source languages datasets $ \{S_i\}_{i=1}^{M}$, where $S_i=\{(x_{s_i}^j,y_{s_i}^j)\}_{j=1}^{|S_i|}$,  and
one unlabeled target language dateset  $T=\{(x_t^j)\}_{j=1}^{|T|} $. 
Specifically, $x_{s_i}^j = \{ w_n \}_{n=0}^{L}$ denotes the $j$-th sentence, with $w_n$ indicating the $n$-th token and $L$ representing  the sentence length.
For RRC and TEP  tasks,  $y_{s_i}^j$ stands for the sentence-level label. For NER, $y_{s_i}^j=\{ y_n^j \}_{n=0}^{L}$ is a label sequence, where  $y_n^j$ specifies  the entity class
corresponding to token $w_n$. In this paper, we aim to train a model with $ \{S_i\}_{i=1}^{M}$ and $T$, and expect it to  generalise well in the target language $T$.

\subsection{Framework and Basic Pipeline}

\subsubsection {Model Framework }  As shown in Figure~\ref{fig:model}, DA-Net consists of a shared encoder and multiple source branch networks.
We  train each source branch  with one source-target language pair to encode language-specific knowledge separately, 
and ultimately, the predictions of target samples can be derived by weighting the outputs of multiple language-specific classifiers. 
To alleviate the semantic interference from multiple sources,  we propose a feedback-guided collaborative disentanglement (FCD) method to purify  input representations. In addition, to mitigate the language gap of language pairs,  we propose a class-aware parallel adaptation (CPA) method to align  class-level distributions across languages. In detail, our  DA-Net has  the following components. Note that we use the RRC task as an example to present  model details throughout the paper.

\subsubsection{Shared Multilingual Encoder}  We adopt mBERT~\cite{DBLP:conf/acl/PiresSG19} as the feature extractor $\bm{E}$ to obtain semantic  representations for different languages. Formally, given one source language sequence, for instance, $x_{s_i}^j$, $\bm{E}$ maps it to a shared latent space and produces  representations $h_{s_i}^j =\bm{E}({x_{s_i}^j};\Psi)$, where $\Psi$  is the encoder parameters.

Since all source languages share the same parameters $\Psi$, which are in turn updated by all these languages,    semantic representations  of each source  language extracted by $\bm{E}$ inevitably contain information from  other source languages.

\subsubsection{Disentangler}  To purify  semantic representations from the shared $\bm{E}$, we equip each source language $S_i$ with a  disentangler $D_i$,  one layer MLP and an activation function.
Taking   source representations $h_{s_i}^j$  as input,  the  $D_{i}$  can produce  new  representations $z_{s_i}^j = D_{i}(h_{s_i}^j; \phi_i)$, where $\phi_i$ is  the  parameters of  $D_{i}$.
All  disentanglers' parameters can be expressed as $\Phi =\{\phi_i\}_{i=1}^M$.
Then, all these disentanglers, guided  by our FCD method, can optimise new  representations and disentangle multiple sources.

\subsubsection{Adaptor} 
To bridge the source-target  language gap, we construct one adaptor $A_{s_i}$  for each language $ S_i $ . The adapter is a one-layer MLP, followed by an activation function. We pass  language representations $z_{s_i}^j$  through the corresponding adaptor and   obtain  $e_{s_i}^j=A_{i}(z_{s_i}^j;\theta_i)$ with  $\theta_i$ denoting the  parameters of $A_{i}$. We express parameters of $M$  adaptors as $\Theta =\{\theta_i\}_{i=1}^M$. 
Then, we perform the CPA method  on $M$  adaptors to boost the adaptation of language pairs.

\subsubsection{Classifier}    Similarly, we construct multiple classifiers  $\{C_{i}\}_{i=0}^{M}$, each is a one-layer MLP, to obtain language-specific predictions. We input representations $e_{s_i}^j$ to  classifier $C_{i}$, and finally obtain the prediction  $p_{s_i}^j =C_{i}( A_{i}(z_{s_i}^j; \omega_i))$, where $\omega_i$  stands for  parameters of $C_{i}$. All $M$ classifiers' parameters  can be expressed as $\Omega =\{\omega_i\}_{i=1}^M$.

With the multiple labelled source languages, our whole model can be
optimized by minimizing cross-entropy loss:
\begin{equation}
\begin{split}
\mathcal L_{CE}(\Psi,\Phi,\Theta,\Omega)
=-\frac{1}{\sum_{i=1}^M |S_i|}   \sum_{i=1}^{M}
 \sum_{j=1}^{|S_i|}   y_{s_i}^j log(p_{s_i}^j )
\end{split}
\label{cewarm}
\end{equation}

\subsection{Feedback-guided Collaborative Disentanglement}
\label{Disentanglement}
\begin{figure}[t]
    \centering
\includegraphics[width=0.48\textwidth]{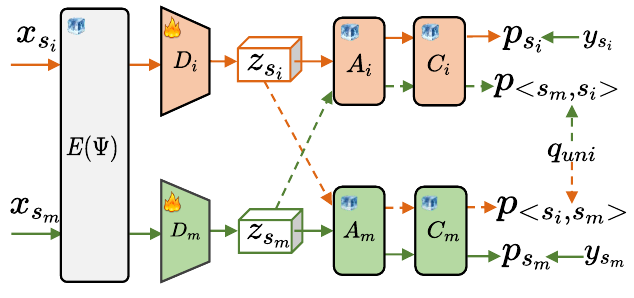}
    \caption{Collaborative disentanglement process. }
    \label{fig:model_two}
\end{figure}

In this part, we detail our  FCD method, which relies on  supervised signals from multiple classifiers to guide  disentanglers to perform a max-min game. In this way, the disentanglers can map each source language into  one exclusive  representation space, respectively, thus eliminating mutual interference among multiple sources.
To illustrate   this process, we take the  source language $S_i$ as  an example (Fig~\ref{fig:model_two}).

\subsubsection{Maximise Prediction Accuracy}
We sequentially feed the  source representation $z_{s_i}^j$ produced by  $E_i$ to adaptor $A_i$ and classifier $ C_i$ (located in source  $S_i$ branch) and obtain prediction $p_{s_i}^j$. To preserve the knowledge of source language  $S_i$, in this step, we   optimize the disentangler  $D_{i}$ to find representations that make the classifier $ C_i$ perform well. To achieve this, we minimize the following cross-entropy loss:

\begin{equation}
\begin{split}
\mathcal L_{max}(\phi_{i})
=-\frac{1}{|S_i|}  
 \sum_{j=1}^{S_i}   y_{s_i}^j log(p_{s_i}^j )
\end{split}
\label{max}
\end{equation}

\subsubsection{Minimise Prediction Accuracy}  In this step, we expect disentanglers to learn   representations on which  other source branches' classifiers  perform poorly, to remove these sources' information. We pass  representations $z_{s_i}^j$ into  adaptors $\{A_{m}\}_{m=1,m\neq i}^{M}$  and classifiers $\{C_{m}\}_{m=1,m\neq i}^{M}$,  and obtain different predictions $\{p_{<s_i,s_m>}^j\}_{m=1,m\neq i}^{M}$. We optimize  disentangler $D_i$ to force each $p_{<s_i,s_m>}^j$ to approximate a uniform distribution $q_{uni}$. Since the uniform distribution presents the highest entropy and  the most randomness \cite{DBLP:conf/eccv/GongLJ20},   the above operation can maximize the prediction chaos  of   classifiers $\{C_{m}\}_{m=1,m\neq i}^{M}$ for  representations $z_{s_i}^j$.
Formally, we achieve the above by minimizing the MSE loss between $p_{<s_i,s_m>}^j$ and $q_{uni}$. Additionally, as different sources have different similarities to each other, undifferentiated  disentanglement may impair the shared information between them. Thus, we employ the distance metric  (i.e. $\alpha_{i,m}$)   as the control factor  to regulate  the disentangling strength among source languages, denoted as: 

\begin{equation}
\begin{split}
\mathcal L_{min}(\phi_{i})
\!=-\frac{1}{\!M|S_i|} \! \!\sum_{m=1, \atop m\neq i}^{\!M} \alpha_{i,m}
 \!\sum_{j=1}^{|S_i|}    ( p_{<s_i,s_m>}^j\!-q_{uni} )^2
\end{split}
\label{min}
\end{equation}

where  $q_{uni}$ is  a ${K}$-dimensional vector with all elements being $\frac{1}{K}$ and  ${K}$ is the number of classes.  $\alpha_{i,m}$ is the MMD distance between $S_i$ and $S_m$ (see Eq \ref{mmd}).

To summarize, the loss $\mathcal L_{max}$ (Eq. \ref{max}) forces disentangler $D_i$  to encode language information from $S_i$, while the loss $\mathcal L_{min}$ (Eq. \ref{min}) discourages  $D_i$  from  encoding information of  $\{S_{m}\}_{m=1,m\neq i}^{M}$.  Consequently, the  information from other sources can be  removed from representations  $z_{s_i}^j$. Eventually, we  can use these two losses to  optimise all disentanglers, and the total loss of this part can be expressed as:

\begin{equation}
\begin{split}
\mathcal L_{FCD}(\Phi)
=\sum_{i=0}^{M} (  \mathcal L_{max}(\phi_{i}) +    \mathcal L_{min}(\phi_i) )
\end{split}
\label{fcd}
\end{equation}

\subsection{Class-aware Parallel  Adaptation}
\label{Adaptation}
In this part, we present  CPA method, which relies on  aligning  class-level distributions  between language pairs on each adaptor to alleviate the language gap.
Here, we utilize the language pair $S_i$ and $T$    as examples to illustrate  the class-level adaptation process.

\subsubsection{Class Pseudo Labels} 

Since there is no annotation in the target language, the class-wise alignment between  source and target  is not trivial. For  unannotated target samples,  an intuitive way is to  employ the training model  to predict their pseudo labels, which may result in unstable predictions. Therefore,  we create a  momentum model $ \mathcal M_{cur}$ derived from the training model instead.
Since $ \mathcal M_{cur}$ utilizes a moving average technique for step-level updates, it can provide  consistent model representation and stable predictions at each step~\cite{DBLP:conf/cvpr/He0WXG20}.

\begin{equation}
\begin{split}
\mathcal M_{cur}= \gamma *\mathcal M_{cur}+(1-\gamma )*\mathcal M_{cur-1}
\end{split}
\label{proto_moving}
\end{equation}

where $\gamma \in (0,1)$ is the moving  coefficient, $cur$ and $cur-1$   denote the current and  previous moment, respectively.

Then, we feed target samples  to   
 $ \mathcal M_{cur}$ and  obtain $M $ set of probabilities. The pseudo labels of target samples  can be derived via weighted summing $M $ different  probabilities  (Eq. \ref{predict}). The class labels  for  source are their golden labels.

\subsubsection{Class-aware Distribution Alignment} 
After obtaining class labels for source and target samples, we can intuitively utilize the lately prevalent supervised contrastive learning ~\cite{DBLP:conf/nips/KhoslaTWSTIMLK20} to achieve class-level  alignment between language pair.  
However, this method tends to pull all samples of the same class together, making  the learned representations ignore intra-class diversity, which is unfavourable to generalize to the target language~\cite{DBLP:conf/emnlp/MouHWWWWX22}. 
Motivated by the fact that   larger intra-class diversity facilitates transfer learning~\cite{DBLP:conf/iclr/ZhaoWLL21,feng2021rethinking},  we design a distribution-based contrastive learning technique to bridge the gap between languages, and maintain the diversity of intra-class features~\cite{DBLP:journals/tnn/WangLDNCDW23}.

In detail, we assume that  source and target samples belonging to class $k$ obey the distributions $P_{s_i}^k$ and $Q_{t}^k$,  respectively. 
We regard one of them as the anchor distribution, such as $P_{s_i}^k$, the other as the positive distribution, such as $Q_{t}^k$, and the remaining class-wise  distributions $\{P_{s_i}^o\}_{o=1,o \neq k}^{K}$ (from source) and $\{Q_{t}^o\}_{o=1,o \neq k}^{K}$  (from target) as negative distributions, denoted as $P_{s_i,neg}^k$. Our goal is to minimise  the distance of  same-class distributions while maximising  the distance of  different-class distributions, denoted as:

\begin{equation}
\begin{split} 
& { \mathcal L_{CA}}(\Psi, \theta_i, \omega_i) = -\sum_{k=1}^{K} \log 
 \\& 
{ \frac{2 e (d(  \!P_{s_i}^k ,  \!Q_t^k) /  \tau) } {  \!\sum_{neg}  \!e( d(  \!P_{s_i}^k,  \!P_{s_i,{neg}}^k) /   \!\tau  ) + \!\sum_{neg  }  \!e ( d(  \!Q_t^k,   \!Q_{t,{neg}}^k) /   \!\tau)} }
\end{split}
\label{duibi}
\end{equation}
where $e(\cdot)$ denotes  exponential function $exp(\cdot)$. $Q_{t,neg}^k$ denotes  negative distributions of  $Q_{t}^k$. 
$\tau$ is a temperature factor.
$d(\cdot)$ is a   distance function  between  language distributions.

As shown, the intra-class distribution discrepancy (numerator) is minimized to align  language distributions within a class. In contrast, the inter-class distribution discrepancy (denominator)  is maximized to push the distributions of different classes  away from  decision boundaries. 
These two   are jointly optimized to improve the adaptation performance.

\subsubsection{Distribution  Metric}  

To optimise  distances of  different class-wise distributions, we suggest using  maximum mean discrepancy (MMD)~\cite{DBLP:conf/ismb/BorgwardtGRKSS06}  metric  to measure  these distances ($d(\cdot)$ in Eq.\ref{duibi} ). MMD is an effective non-parametric metric  for comparing two distributions
based on  two sets of observed samples belonging to them~\cite{DBLP:journals/jmlr/GrettonBRSS12}.
Suppose we have two class-wise distributions $P_{s_i}^u$,
 $Q_{t}^{v}$ and their respective observations samples $\{x_{u}^j\}_{j=1}^{|S_i^u|}$, $\{x_{v}^j\}_{j=1}^{|T^v|}$, with  $u$, $v$ denoting different classes. Through introducing kernel tricks and empirical kernel mean embeddings ~\cite{DBLP:journals/jmlr/GrettonBRSS12},  the empirical estimation of MMD  between  $P_{s_i}^u$ and
 $Q_{t}^{v}$ is expressed as:

\begin{table*}[]
\renewcommand\arraystretch{1.0}
\setlength\tabcolsep{2.0pt}
\centering
\begin{tabular}{ccccccccccccccccc}
\toprule[1.5pt]
\textbf{Methods} &
  \textbf{af} &
  \textbf{hr} &
  \textbf{it} &
  \textbf{fr} &
  \textbf{fa} &
  \textbf{ur} &
  \textbf{zh-yue} &
  \textbf{he} &
  \textbf{id} &
  \textbf{az} &
  \textbf{ja} &
  \textbf{et} &
  \textbf{eo} &
  \textbf{no} &
  \textbf{sh} &
  \textbf{Avg} \\ \hline
\multicolumn{17}{c}{Source Language: en, es and ru} \\ \hline
mBERT &
  {  81.33} &
  83.99 &
  82.83 &
  {  85.57} &
  56.93 &
  38.69 &
  49.45 &
  60.93 &
  66.47 &
  72.82 &
  36.35 &
  78.81 &
  {  68.84} &
  83.13 &
  81.55 &
  68.51 \\
M-MOE &
  80.41 &
  82.26 &
  83.99 &
  83.49 &
  55.00 &
  46.54 &
  {  51.35} &
  \textbf{63.34} &
  64.04 &
  70.75 &
  {  37.25} &
  77.44 &
  68.75 &
  82.35 &
  82.05 &
  68.60 \\
MAN &
  80.22 &
  84.01 &
  83.14 &
  84.95 &
  57.41 &
  {  49.34} &
  48.65 &
  60.38 &
  64.69 &
  72.47 &
  36.73 &
  80.00 &
  65.82 &
  84.30 &
  79.03 &
  68.74 \\
MulTS &
  81.28 &
  84.26 &
  83.44 &
  84.22 &
  53.20 &
  42.32 &
  48.82 &
  61.43 &
  66.68 &
  {  73.73} &
  35.01 &
  {  82.87} &
  67.25 &
  84.15 &
  79.21 &
  68.52 \\
G-MOE &
  79.12 &
  {  85.39} &
  {  84.38} &
  85.12 &
  {  58.16} &
  49.71 &
  49.55 &
  62.58 &
  {  70.36} &
  73.31 &
  33.59 &
  80.97 &
  66.39 &
  {  86.00} &
  {  82.82} &
  {  69.83} \\
DA-Net &
  \textbf{83.57} &
  \textbf{86.24} &
  \textbf{85.02} &
  \textbf{87.16} &
  \textbf{66.08} &
  \textbf{56.79} &
  \textbf{52.44} &
  {  61.84} &
  \textbf{72.51} &
  \textbf{74.52} &
  \textbf{38.38} &
  \textbf{83.65} &
  \textbf{70.19} &
  \textbf{86.58} &
  \textbf{87.50} &
  \textbf{72.83} \\ \hline
\multicolumn{17}{c}{Source Language: en, ar and zh} \\ \hline
mBERT &
  77.31 &
  79.81 &
  78.28 &
  81.37 &
  65.74 &
  37.26 &
  68.06 &
  61.75 &
  67.33 &
  69.30 &
  46.08 &
  78.62 &
  60.59 &
  77.92 &
  56.24 &
  67.04 \\
M-MOE &
  77.58 &
  78.88 &
  82.33 &
  79.75 &
  {  65.87} &
  48.24 &
  69.86 &
  {63.32} &
  64.43 &
  69.83 &
  43.86 &
  79.49 &
  {  62.49} &
  80.78 &
  58.57 &
  68.28 \\
MAN &
  78.17 &
  80.19 &
  81.07 &
  80.67 &
  65.34 &
  46.33 &
  70.14 &
  61.89 &
  64.04 &
  70.76 &
  42.64 &
  79.84 &
  61.50 &
  78.01 &
  56.91 &
  67.81 \\
MulTS &
  {  80.33} &
  81.92 &
  81.78 &
  82.85 &
  53.87 &
  {  52.44} &
  65.86 &
  63.01 &
  58.96 &
  71.85 &
  43.22 &
  \textbf{  82.90} &
  61.49 &
  {  82.03} &
  60.50 &
  68.20 \\
G-MOE &
  78.64 &
  {  82.29} &
  {  82.88} &
  {  83.13} &
  62.17 &
  52.07 &
  {  70.26} &
  {  63.61} &
  {  74.33} &
  {  73.47} &
  {  47.64} &
  81.15 &
  60.80 &
  80.69 &
  {  72.50} &
  {  71.04} \\
DA-Net &
  \textbf{81.46} &
  \textbf{83.46} &
  \textbf{83.20} &
  \textbf{83.91} &
  \textbf{80.81} &
  \textbf{69.63} &
  \textbf{72.42} &
  \textbf{63.73} &
  \textbf{81.27} &
  \textbf{74.39} &
  \textbf{49.57} &
  81.83 &
  \textbf{62.53} &
  \textbf{83.53} &
  \textbf{83.34} &
  \textbf{75.67} \\ \toprule[1.5pt]
\end{tabular}
    \caption{ The   NER task results  on the Wikiann dataset. }
\label{tab:NER_result}
\end{table*}

\begin{equation}
\begin{aligned}
\mathcal{D}_{\mathrm{MMD}}^{2}(P_{s_i}^{u}, Q_{t}^{v}) & =\frac{1}{\left|\boldsymbol S_i^{u}\right|\left|\boldsymbol S_i^{u}\right|} \sum_{j=1}^{\left|\mathcal S_i^{u} \right|} \sum_{l=1}^{\left|\mathcal S_i^{u} \right|} G\left(e^j_{u}, e_{u}^l\right) \\
& +\frac{1}{\left|\boldsymbol T^{v} \right|\left|\boldsymbol T^{v} \right|} \sum_{j=1}^{\left|\mathcal T^{v} \right|} \sum_{l=1}^{\left|\mathcal T^{v} \right|} G\left(e_{v}^j, e_{v}^l\right) \\
& -\frac{2}{\left|\boldsymbol S_i^{u} \right|\left|\boldsymbol T^{v} \right|} \sum_{j=1}^{\left|\mathcal S_i^{u} \right|} \sum_{l=1}^{\left|\mathcal T^{v} \right|} G\left(e_{u}^j, e_{v}^l\right)
\end{aligned}
\label{mmd}
\end{equation}

where $e_{u}^j$ and  $e_{v}^j$ are representations  corresponding to samples $x_{u}^j$ and $x_{v}^j$, respectively,  generated by  adaptor $A_i$. Note that, $x_{u}^j \in S_i^u$ (class $u$ of source  $S_i$), $x_{v}^j \in T^v$ (class $v$ of target language $T$). $G$ refers to the Gaussian kernel.

We  can employ the MMD metric to  calculate   distances between each pair of class-wise distributions in Eq \ref{duibi}, regardless of  whether these distributions originate  from source or target, and belong to the same class or not. In addition, the MMD can also be utilized to derive the value of  $\alpha_{i,m}$ (Eq \ref{min}). In this case,  we adopt samples from $S_i$ and $S_m$ to estimate the  distance between language distributions $P_{s_i}$ and $P_{s_m}$.

  Subsequently, we will implement class-level distribution alignment on each branch via $\mathcal L_{CA}$. Eventually, our  class-aware parallel adaptation loss can be  defined as follows:

 \begin{equation}
\begin{split} 
{ \mathcal L_{CPA}(\Psi, \Theta,\Omega)} = \sum_{i =0}^{M} \mathcal L_{CA} (\Psi, \theta_i,\omega_i)
\label{CPA}
\end{split}
\end{equation}

\subsection{Training and Prediction}
\label{training and  prediction}

We first describe the training procedure, followed by  the target prediction of our model.

\subsubsection {Trainning Procedure}
To equip our model with initial classification performance to serve as  language detectors in  disentangling stage and to acquire pseudo labels in  adaptation  stage, firstly, we warm up our model via   $\mathcal L_{CE}$ (Eq \ref{cewarm}) with   multiple labelled source languages   in the first epoch.

Then, we  iteratively perform   disentangling and adaptation  via $\mathcal L_{FCD}$ (Eq \ref{fcd}) and $\mathcal L_{CPA}$(Eq \ref{CPA}) at the  batch level. 
In the disentangling stage,   we fix all the parts except disentanglers, and minimize the loss $\mathcal L_{FCD}$  to  update these disentanglers. In the adaptation step,  we fix all the disentanglers and update the  other components of  the model  by minimizing the losses $ \mathcal L_{CE} + \eta \mathcal L_{CPA} $, with $\eta$ denoting the trade-off  factor.

\subsubsection{Target Prediction} 
We use $x_t^j$ as an example to describe the prediction process. Firstly, we  input this sample into our model and obtain multiple predictions $\{p_{<t,s_i>}^j\}_{i=1}^{M}$ from  language-specific classifiers. 
Then, all $M$ predictions are weighted  via a point-to-set Mahalanobis distance-based  metric ~\cite{mclachlan1999mahalanobis}, which measures the similarity between  target sample $x_t^j$ and each source.

\begin{equation}
\begin{split}
y_t^j=\sum_{i=1}^{M} \beta (x_t^j, S_i) p_{<t,s_i>}^j
\end{split}
\label{predict}
\end{equation}
 where $\beta(x_t^j, S_i) =-((x_t^j-\mu_{s_i})^T\Sigma_{s_i}^{-1}(x_t^j-\mu_{s_i}))^{\frac{1}{2}} $. The $\mu_{s_i}$ and $\Sigma_{s_i}^{-1}$ are the mean encoding and  the inverse covariance matrix of $S_i$, respectively.

\begin{table*}[]
\renewcommand\arraystretch{1.0}
\setlength\tabcolsep{4.2pt}
\begin{tabular}{cccccccc|ccccccc|c}
\toprule[1.5pt]
\textbf{Methods} & \textbf{de} & \textbf{fr} & \textbf{bg} & \textbf{th} & \textbf{sw} & \textbf{vi} & \textbf{hi} & \textbf{de} & \textbf{fr} & \textbf{bg} & \textbf{th} & \textbf{sw} & \textbf{vi} & \textbf{hi} & \multirow{2}{*}{\textbf{Avg}} \\ \cline{1-15}
\multicolumn{8}{c|}{Source Language: en, es and ru} & \multicolumn{7}{c|}{Source Language: en, ar and zh} &  \\ \cline{1-16}
mBERT & 68.22 & 70.49 & 67.29 & 52.07 & 50.17 & 67.11 & 60.77 & 66.81 & 68.54 & 66.37 & 53.19 & 50.68 & 67.45 & 60.12 & 62.09 \\
M-MOE & 68.90 & 69.92 & 67.96 & 52.91 & 49.76 & 67.13 & 57.28 & 68.42 & 68.80 & 67.18 & 54.55 & 50.71 & 66.26 & 60.79 & 62.18 \\
MAN & 69.04 & 69.15 & 67.94 & 53.19 & 49.82 & 66.94 & 60.65 & 67.48 & 68.90 & 67.94 & 53.29 & 49.82 & 67.32 & 60.65 & 62.30 \\
MulTS & 68.32 & 69.50 & 67.20 & 53.37 & 48.54 & 67.66 & 61.45 & 67.14 & 68.68 & 67.10 & 52.55 & 49.40 & 67.96 & 60.69 & 62.11 \\
G-MOE & 68.89 & 69.71 & 68.67 & 53.59 & 50.73 & 66.33 & 61.46 & 67.63 & 69.77 & 67.59 & 52.96 & 49.16 & 68.35 & 59.80 & 62.62 \\
DA-Net & \textbf{69.96} & \textbf{71.54} & \textbf{69.18} & \textbf{55.29} & \textbf{51.16} & \textbf{68.49} & \textbf{61.47} & \textbf{69.94} & \textbf{70.98} & \textbf{68.88} & \textbf{55.52} & \textbf{51.71} & \textbf{68.96} & \textbf{61.52} & \textbf{63.90} \\ \toprule[1.5pt]
\end{tabular}
 \caption{ The  TEP task results  on the XNLI dataset.}
 \label{tab:tep_result}
\end{table*}

\begin{table}[]
\renewcommand\arraystretch{1.0}
\setlength\tabcolsep{3.0pt}
  \centering
\begin{tabular}{cccc|ccc|c}
 \toprule[1.5pt]
\textbf{Source} & \multicolumn{3}{c|}{\textbf{en,es, fr}} & \multicolumn{3}{c|}{\textbf{en, ja, zh}} & \multirow{2}{*}{\textbf{Avg}} \\ \cline{1-7}
\textbf{Methods} & de             & ja    & zh    & de             & fr    & es    &       \\ \hline
mBERT            & 50.08          & 40.46 & 40.77 & 49.46          & 50.74 & 50.42 & 46.99 \\
M-MOE            & 50.94 & 40.86 & 40.92 & 49.69          & 51.94 & 49.28 & 47.27 \\
MAN              & 50.19          & 40.94 & 41.42 & 48.84          & 50.16 & 51.33 & 47.15 \\
MulTS            & 49.88          & 39.41 & 40.25 & \textbf{49.98} & 49.34 & 50.02 & 46.48 \\
G-MOE            & 49.90          & 41.36 & 41.19 & 49.75          & 51.18 & 50.34 & 47.29 \\
DA-Net          & \textbf{50.99} & \textbf{42.48} & \textbf{42.54} & 49.84  & \textbf{53.58} & \textbf{52.24} & \textbf{48.61}                \\  \toprule[1.5pt]
\end{tabular}
    \caption{The   RRC task results  on the Amazon dataset.}
\label{tab:text_classification}
\end{table}


\section{Experiments and Analysis}
\subsection{Experiment Setting}

 \subsubsection{Datasets} 

For NER task, 
we employ the  adopt the benchmark  dataset Wikiann~\footnote{https://huggingface.co/datasets/wikiann}~\cite{DBLP:conf/acl/RahimiLC19}, where  each word  is  marked by the BIO scheme, and  annotated with  LOC (Location), PER (Person), or ORG (Organisation).

For  RRC task, we adopt  the benchmark  Amazon Reviews 
Corpus~\footnote{https://huggingface.co/datasets/amazon\_reviews\_multi}~\cite{DBLP:conf/emnlp/KeungLSS20} dataset, where  each customer review is classified into five sentiment star ratings (classes).

For  TEP  task, we adopt  the benchmark dataset XNLI~\footnote{https://huggingface.co/datasets/xnli}~\cite{DBLP:conf/emnlp/KeungLSS20},  where  each sentence pair is labeled with entailment, neutral or contradiction. Due to the  computational resources limitation, we  take $20,000$ samples on each class for each language separately as   training set.


Our experiments  cover $38$ languages and  contain  $4$ low-resource languages,  namely Esperanto (eo), Norwegian (no), Serbo-Croatian (sh), and Cantonese (zh-yue). These languages are not involved at all during the pre-trained model mBERT pretraining. For all datasets, we adopt the original  training, development, and evaluation sets. Note that, the source and target data of the Amazon Reviews 
Corpus and XNLI  share the same label distribution, while those of Wikiann  have different label distributions, as detailed in the given dataset addresses.
To simulate the zero-shot scenario, we train the model with both the labelled source and unlabeled target  training sets, validate the model on the source  validation set, and evaluate the model on the target  test set.

\subsubsection{Implementation Details} 
Following previous works~\cite{DBLP:conf/emnlp/KeungLSS20, DBLP:conf/emnlp/MaCGLGL0L22}, we adopt token-level F1 metric  for NER task and accuracy metric for RRC and TEP tasks. 
We train our model using Adam~\cite{DBLP:journals/corr/KingmaB14} optimizer.  We set the batch size to $128$ for NER  and $64$ for  other tasks. We use  maximum sequence length =  $128$,  the moving  coefficient $\gamma$ = $0.0001$, and the dropout =  $0.5$  empirically.  We utilize the grid search technology to obtain the
optimal super-parameters, including the temperature factor $\tau$ selected from $\{0.1, 0.3,0.5,0.7, 0.9\}$, the trade-off factor $\eta$  selected from $\{0.01,0.05,0.1,0.5\}$, the learning rate for encoder selected from $\{1e-5, 3e-5, 5e-5\}$ and for other components selected from $\{0.0001,0.0005\}$. 
We repeat all experiments $5$ times   and report their mean results.All experiments are implemented using PyTorch and are conducted on  NVIDIA Tesla V100 GPU.

\subsection{Experimental Results and Aanlysis}

\subsubsection{Baseline and SOTAs} 
We compare our DA-Net with several previous multi-source approaches.   \textbf{ (1) mBERT}~\cite{DBLP:conf/acl/PiresSG19}, \textbf{ (2) M-MOE}~\cite{DBLP:conf/emnlp/GuoSB18},
 \textbf{ (3) MAN}~\cite{DBLP:conf/acl/ChenAHWC19}, \textbf{ (4) MulTS}~\cite{DBLP:conf/acl/WuLKLH20} and \textbf {(5) G-MOE}~\cite{DBLP:conf/acl/JinD0LCDQ22}.  Note that the original M-MOE and MAN use LSTM as the encoder. For a fair comparison, we replace the encoder with mBERT.

\subsubsection{Performance Comparison}

To  evaluate the model performance, we conduct experiments using two distinct source combinations for each  task: the same  language family  (Setting 1) and different   language families  (Setting 2). For example, in NER, sources en, es, and ru  belong to the same language family (Indo-European), while en, ar, and zh belong to the Indo-European, Afro-Asian, and Sino-Tibetan, respectively. 
As shown in  Table \ref{tab:NER_result}, \ref{tab:tep_result} and \ref{tab:text_classification}, our method outperforms  the baseline and the SOTAs in almost  all target languages for three tasks under different settings. In detail, for NER, we exceed the SOTA G-MOE by 3\% points in  Setting 1,and 4.63\% in Setting 2 on average. Since our method employs   FCD and CPA strategies in training,  we can simultaneously reduce interference among  sources and enhance source-target languages' adaptation.

\begin{figure}[t]
    \centering
\includegraphics[width=0.48\textwidth]{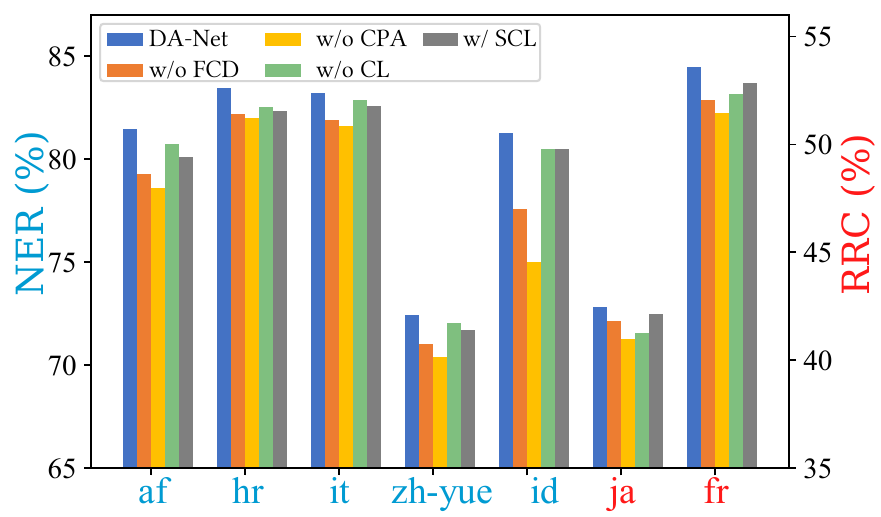}
    \caption{Ablation Study.}
    \label{fig:T-NSE-NER}
\end{figure}

\begin{figure*}[t]
    \centering
\includegraphics[width=0.95\textwidth]{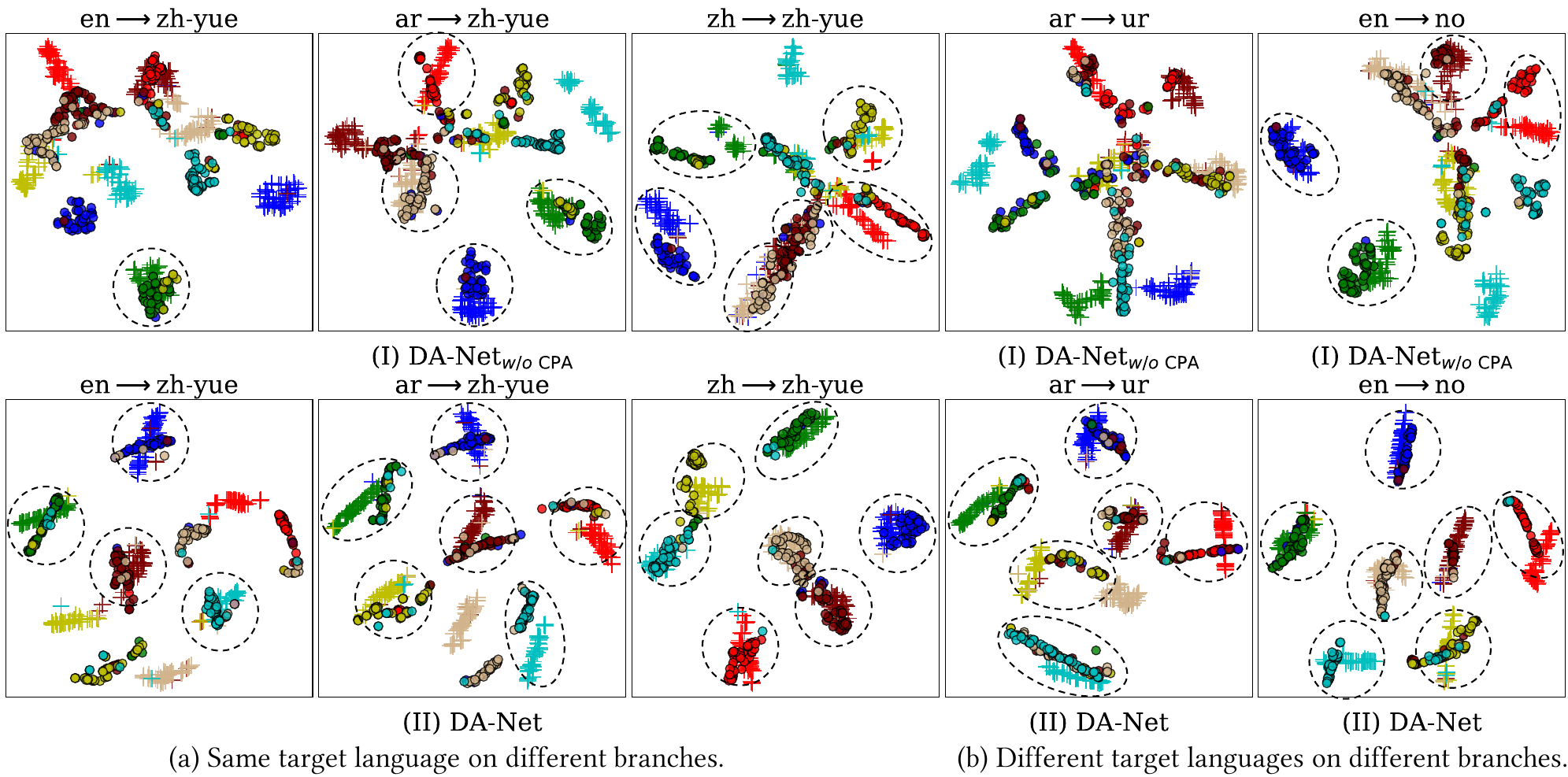}
    \caption{Our method achieves more class-level alignment and learns better class-discriminative representations. The plus  ($+$) and circles
($\bullet$) indicate  representations of the source and target languages. Different colours represent different classes.}
    \label{fig:T-NSE-NER}
\end{figure*}
 
More specifically, in Setting 2, our model is 18.64\% and 17.56\%  higher than SOTA  for  fa and ur languages. While both  belong to the Indo-European language family, they are written using the Arabic alphabet. Benefiting from our CPA method, we can  preserve the shared knowledge of each language pair. Thus, these target languages can transfer  grammatical and semantic knowledge from en and script knowledge from ar. In contrast, the SOTA model forces all languages to be aligned, compromising the exclusive knowledge of the language pairs.
In addition, our method also performs well when the target language is outside the range of the sources' language family, e.g., id (Austronesian  language family), ja (Ryukyu language family). 
On low-resource languages such as zh-yue, eo, no, sh, our model still surpasses  SOTA, demonstrating prominent generalisation.

\begin{figure}[h]
    \centering
\includegraphics[width=0.48\textwidth]{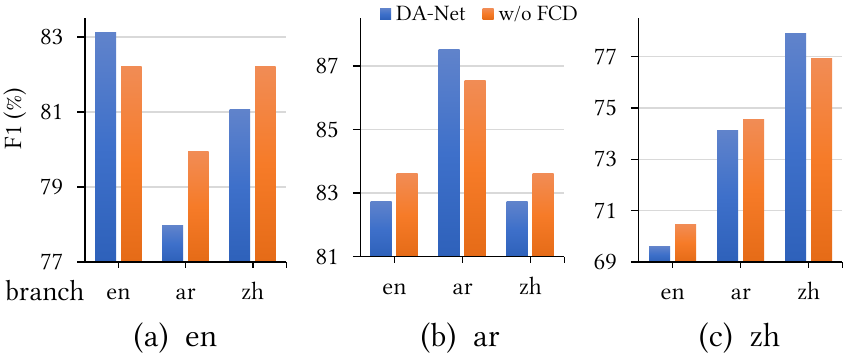}
    \caption{Collaborative Disentanglement Analysis.}
    \label{fig:decoupling}
\end{figure}

\subsubsection{Ablation Study} 

To explore the contribution of each component, we design five variant models. (1) DA-Net$_{w/o \;\rm{FCD}}$  removes the FCD method. (2) DA-Net$_{w/o \;\rm{CPA}}$  wipes out the CPA method. (3) DA-Net$_{w/o \;\rm{CL}}$ employs  MMD loss between language distributions  instead of class-aware  MMD loss in the CPA method. (4) DA-Net$_{w/ \;\rm{SCL}}$  replaces the distributional alignment to supervised contrast learning in the CPA method. 
As shown in  Fig 4, for NER, compared to DA-Net, the af of DA-Net$_{w/o \;\rm{FCD}}$ decreases by  2.16\%. This suggests that the FCD method facilitates performance improvement by alleviating the language interference issue.  DA-Net$_{w/o \;\rm{CPA}}$ decreased by 6.25\% on id, indicating that CPA method can effectively improve model generalisation since it bridges the gap of language pairs. 
The slight drop of 
DA-Net$_{w/ \;\rm{SCL}}$ and  DA-Net$_{w/o \;\rm{CL}}$    indicates that  the distribution alignment and  class information  are helpful for final  performance.  Similar results can be observed on RRC task.

\subsubsection{Collaborative Disentanglement Analysis} 

To demonstrate that the FCD approach can mitigate interference between multiple sources, we compare the performance of DA-Net and DA-Net$_{w/o \;\rm{FCD}}$ in the NER task in Figure~\ref{fig:decoupling}. We input three source languages (en, ar and zh) into the three branch networks of the two models, and get prediction results. For example, Figure~\ref{fig:decoupling} (a) is the results of en in three branches. As shown,   DA-Net improves the F1 of each source language on the corresponding branch and decreases its F1 on the other branches.   This indicates that  DA-Net can purify representations from the shared encoder,  keeping language-specific classifiers from interfering with each other during training, and improving the performance of language-specific classifiers for  their corresponding source.

\subsubsection{Visualize Representations} 
To present that the CPA method can achieve class-level alignment in each branch, we compare the representations produced by adaptors of DA-Net and DA-Net$_{w/o \;\rm{CPA}}$.
Fig~\ref{fig:T-NSE-NER} (a) visualises representations of target zh-yue and three different sources (en, ar, zh), while Fig ~\ref{fig:T-NSE-NER} (b) visualises representations of other two target languages and different sources. Due to the language gap,  the feature distributions of source and target language produced by DA-Net$_{w/o \;\rm{CPA}}$ are significantly different. Many target language examples of one class are incorrectly aligned to source examples of a different class, resulting in poor adaptation. In contrast,  DA-Net correctly aligns more class-level distributions and captures more shared knowledge across languages. More importantly, DA-Net successfully separates and pulls apart clusters of different classes, learning better class-differentiated representations.

\section{Conclusion}

This paper presents a model DA-Net for multi-source cross-lingual tasks. DA-Net proposes a FCD approach to purify input representations of the classifier, thus mitigating inter-multisource interference and optimizing the learning of language-specific classifiers. In addition, DA-Net introduces a CPA method to improve language pairs’ adaption, thus enhancing the performance of language-specific classifiers for target languages.
Experimental results demonstrate DA-Net is effective and outperforms previous SOTAs.


\section{Acknowledgments}
We sincerely thank the reviewers for their insightful comments and valuable suggestions. This work is funded by the National Key Research and Development Program of the Ministry of Science and Technology of China (No. 2021YFB1716201). Thanks for the computing infrastructure provided by Beijing Advanced Innovation Center for Big Data and Brain Computing.

\bibliography{aaai24}

\end{document}